\documentclass[journal]{IEEEtran}

\usepackage[utf8]{inputenc}
\usepackage[T1]{fontenc}
\usepackage{amsmath}
\usepackage{amsfonts}
\usepackage{amssymb}
\usepackage{graphicx}
\usepackage{booktabs}
\usepackage{caption}
\usepackage{subcaption}
\usepackage{times}
\usepackage{url}
\usepackage{hyperref}
\usepackage{bm}
\usepackage{xcolor}
\usepackage{array, multirow}
\usepackage{makecell}
\usepackage{pifont}
\usepackage{siunitx}
\usepackage[capitalise]{cleveref}
\usepackage{wrapfig}

\newcounter{task}

\makeatletter
\newcommand{\teasercaption}[1]{%
  \refstepcounter{figure}%
  \@makecaption{\fnum@figure}{#1}%
}
\makeatother

\newcolumntype{C}{S[table-format=2.1(2.1), uncertainty-mode=separate, table-align-uncertainty=true]}
\newcolumntype{B}{>{\bfseries}S[table-format=2.1(2.1), uncertainty-mode=separate, table-align-uncertainty=true]}
\newcolumntype{D}{S[table-format=1.2(1.2), uncertainty-mode=separate, table-align-uncertainty=true]}
\newcolumntype{E}{>{\bfseries}S[table-format=1.2(1.2), uncertainty-mode=separate, table-align-uncertainty=true]}
\newcolumntype{F}{S[table-format=1.3(1.3), uncertainty-mode=separate, table-align-uncertainty=true]}
\newcolumntype{G}{>{\bfseries}S[table-format=1.3(1.3), uncertainty-mode=separate, table-align-uncertainty=true]}

\usepackage{titlesec}
\titlespacing{\subsection}{0pt}{6pt}{3pt}
\titlespacing{\section}{0pt}{12pt}{6pt}

\begin{document}

\markboth{IEEE Robotics and Automation Letters. Preprint Version. Accepted September, 2026}
{Dong \MakeLowercase{\textit{et al.}}: SemAnCorr} 

\title{\LARGE \bf \emph{SemAnCorr}: \emph{Sem}antic \emph{An}chored \emph{Corr}espondence \\ for Zero-Shot Manipulation Skill Transfer
} 

\author{Xiaoxiang~Dong$^{1}$, William~Baron$^{1,2}$, Hongyi~Chen$^{1}$, Uksang~Yoo$^{1,3}$,
Jeffrey~Ichnowski$^{1}$, Weiming~Zhi$^{2,4}$%
    \thanks{Manuscript received: July 27, 2026; Accepted: September 26, 2026.}%
    \thanks{This paper was recommended for publication by Editor Wei Pan upon evaluation of the Associate Editor and Reviewers' comments.}%
    \thanks{$^{1}$Robotics Institute, Carnegie Mellon University, USA.}%
    \thanks{$^{2}$College of Connected Computing, Vanderbilt University, USA.}%
    \thanks{$^{3}$College of Engineering, University of California, Berkeley, USA.}%
    \thanks{$^{4}$School of Computer Science, The University of Sydney, Australia.}%
    \thanks{Digital Object Identifier (DOI): see top of this page.}%
}

\IEEEaftertitletext{
    \vspace{-2em}
    \begin{center}
    \includegraphics[width=0.9\textwidth]{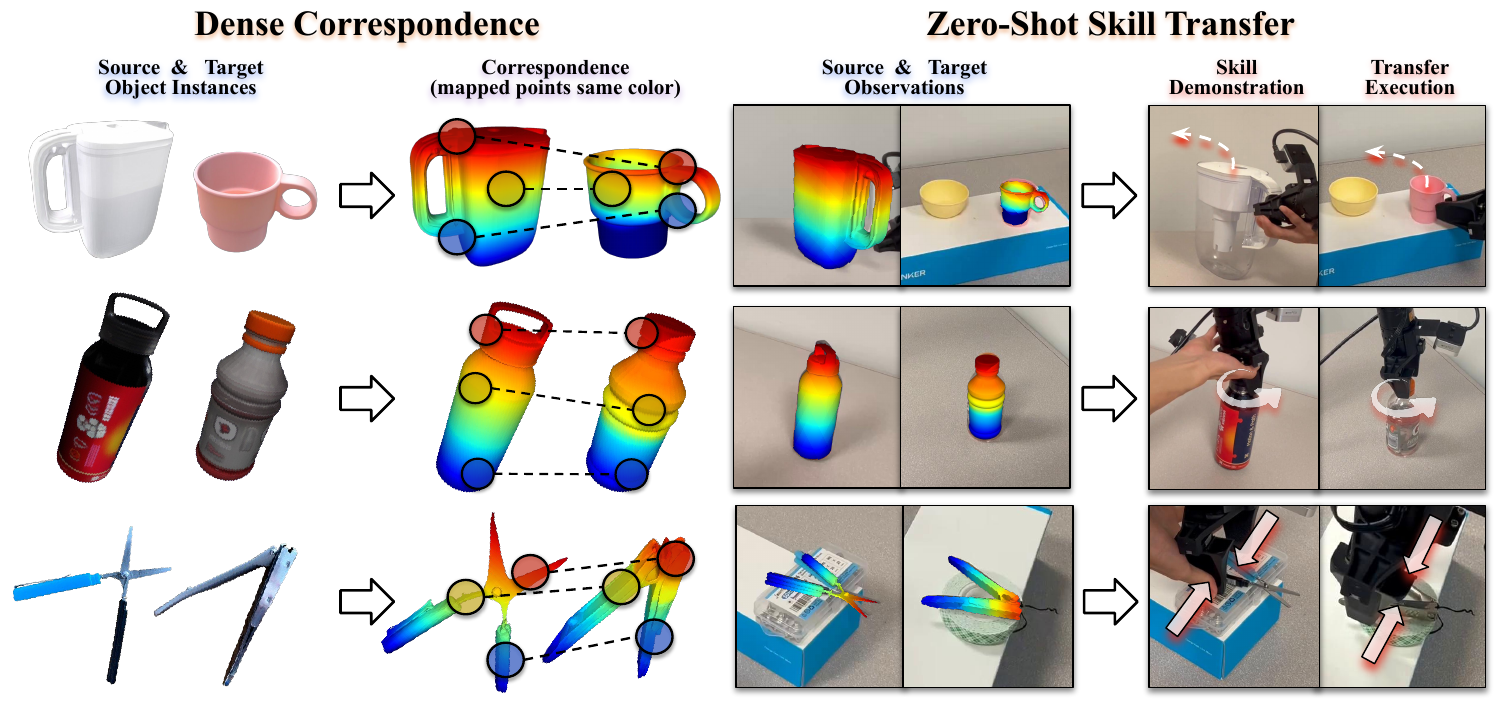}
    \teasercaption{We propose \textbf{SemAnCorr}, a dense vertex-level correspondence across geometrically diverse object instances that respects semantically similar parts between objects while smoothly spanning object surface (left). The resulting correspondence enables zero-shot transfer of manipulation skills  across instances that share functional structure but differ in geometry (right).}
    \label{fig:figure1}
    \vspace{-0.5em}
    \end{center}
}

\maketitle

\begin{abstract}
Transferring manipulation skills across object instances that share functionality but differ in geometry remains a fundamental challenge in robot learning. While recent correspondence methods leverage dense visual descriptors and 3D feature fields, nearest-neighbor feature matching often produces spatially incoherent correspondences that fail to recover the local geometric frames required for reliable skill transfer. We introduce SemAnCorr, a training-free framework that establishes dense correspondence by selecting semantically consistent anchor regions through joint pose-correspondence optimization and propagating these constraints over the object surface using functional maps. The resulting correspondences preserve both semantic consistency and geometric coherence, enabling object-centric manipulation skills to transfer across geometrically diverse instances. We evaluate SemAnCorr on a dense correspondence benchmark built on PartNet-Mobility, achieving 90.8\% semantic accuracy in our benchmark evaluation while improving geometric coherence over recent state-of-the-art baselines. Finally, we show that these improvements translate directly into real-world manipulation performance:  compared to existing correspondence methods, SemAnCorr enables substantially more reliable zero-shot manipulation skill transfer from a demonstration to an unseen object across 5 real-world tasks. Videos and additional visualizations are available at \href{https://semancorr.github.io/}{semancorr.github.io}.
\end{abstract}

\begin{IEEEkeywords}
Perception for Grasping and Manipulation; Learning from Demonstration; Transfer Learning
\end{IEEEkeywords}

\vspace{-1.0em}

\section{Introduction}

\IEEEPARstart{R}{ecent} advances in imitation learning enable robots to acquire manipulation skills from only a few demonstrations~\cite{dong2026jfto, chen2026dexterousmanipulationpoliciesrgb, tamura2026skillweaveweavingheterogeneousdemonstrations}, but the acquired skills typically remain tied to the object on which they were demonstrated. Ideally, a robot should be able to reuse such a skill on a new object that serves the same function, even when its geometry is substantially different: a mug with a different handle shape, a bottle with a different cap, and a pair of pliers with a different jaw length. 

Existing approaches typically transfer demonstrations by identifying semantically corresponding interaction regions or affordances. While sufficient for locating approximate contact regions, these methods often fail to capture the functional intent of manipulation skills: not just where to interact with an object, but in what local frame relative to the object's structure. 

Dense correspondence naturally addresses this limitation by preserving a dense mapping between object surfaces. For manipulation, semantic correspondence determines \textbf{where} a robot should interact, whereas geometrically coherent correspondence recovers the \textbf{local frame} in which that interaction is executed. Effective skill transfer therefore requires correspondences that are both semantically consistent and geometrically coherent, motivating a dense correspondence formulation.

We propose \emph{Sem}antic \emph{An}chored \emph{Corr}espondence (SemAnCorr), a training-free framework that establishes dense correspondence across object instances by anchoring semantically meaningful regions and propagating these constraints over the object surface using functional maps (\cref{fig:figure1}). By jointly preserving semantic consistency and geometric coherence, SemAnCorr recovers the local geometric structure required to transfer demonstrated manipulation skills across geometrically diverse objects.

We evaluate SemAnCorr on a dense correspondence benchmark constructed from PartNet-Mobility~\cite{Xiang_2020_SAPIEN} and validate its practical utility through real-world manipulation. Experiments show that improved geometric correspondence translates directly into more reliable zero-shot skill transfer on real-world tasks.

In summary, this paper makes the following contributions:
\begin{itemize}
    \item We introduce \textbf{SemAnCorr}, a training-free dense correspondence framework that combines semantic anchor selection with functional map propagation to produce correspondences that are both semantically consistent and geometrically coherent.
    \item We construct a \textbf{dense correspondence benchmark} on PartNet-Mobility that evaluates both semantic accuracy (\emph{where} to interact) and geometric coherence (\emph{local frame} to execute the interaction).
    \item We develop an \textbf{object-centric manipulation skill transfer pipeline} that leverages SemAnCorr to transfer demonstrated skills across object instances, enabling zero-shot manipulation from a single demonstration.
\end{itemize}

\section{Related Work}\label{sec:related_work}

\textbf{Dense 3D Correspondence.}
Classical functional map methods~\cite{ovsjanikov2012functional} use geometric descriptors such as WKS~\cite{Aubry2011TheWK} but rely solely on intrinsic geometry and fail across semantically diverse objects. Deep functional map methods~\cite{Donati2020DeepGF} learn feature extractors but require category-specific training data. DenseMatcher~\cite{zhu2024densematcher} combines SD-DINO features with a learned DiffusionNet refiner and functional maps, achieving strong cross-category correspondence. SemAnCorr shares the functional map backbone but replaces learned refinement with semantic anchor selection from pretrained features, enabling training-free generalization at the cost of per-pair optimization.

\textbf{Semantic Correspondence for Manipulation.}
Semantic correspondence has emerged as a principled approach to skill transfer without large-scale demonstration collection~\cite{Florence2018DenseON}. Recent work leverages 2D foundation model features: DIFT~\cite{tang2023emergent} extracts correspondences from diffusion features, SD-DINO~\cite{Zhang2023ATO} fuses Stable Diffusion and DINOv2, and Robo-ABC~\cite{ju2025robo} and RAM~\cite{Kuang2024RAMRA} apply these ideas to affordance transfer and demonstration retrieval, respectively. Vision-language models further broaden this generalization, either as a teacher, as in UAD~\cite{Tang2025UADUA}, which distills VLM-proposed affordances into visual features, or as a planner, as in CrossInstruct~\cite{baron2026crossinstruct}, which synthesizes robot motion from cross-modal instructions over multiple 2D views. Adapting by Analogy~\cite{Gupta2025AdaptingBA} extends functional correspondence to the policy level. While effective, these approaches operate primarily in 2D and neglect 3D quantities, such as local interaction frames, that are crucial for manipulation.

\textbf{3D Feature Fields for Manipulation Generalization.}
Lifting visual features into 3D has proven effective for manipulation generalization~\cite{Florence2018DenseON, Yen2022NeRFSupervision}. D3Fields~\cite{wang2024d3fields} builds dynamic 3D descriptor fields by lifting DINOv2 features from \mbox{RGB-D} observations, and O3Afford~\cite{Tian2025O3AffordO3} combines 3D features with cross-attention for one-shot affordance grounding. These fields capture fine-grained local semantics, but point-wise nearest-neighbor matching can produce spatially incoherent correspondences across distant surface regions; SemAnCorr addresses this with functional map propagation, enforcing global surface smoothness.

\textbf{Affordance Learning and Tool-Use Manipulation.}
Affordance learning for tool use has been approached through actionability prediction~\cite{Mo2021Where2ActFP, Deng20213DAA} and part-centric representations~\cite{Geng2022GAPartNetCD}, establishing which object regions support specific actions, with AdaAfford~\cite{Wang2021AdaAffordLT} extending this to few-shot generalization. However, these methods require category-specific training. Keypoint and retrieval-based methods offer more transferable representations: KETO~\cite{Qin2019KETOLK} learns task-specific keypoints for tool manipulation, GIFT~\cite{Turpin2021GIFTGI} identifies functional affordances without labels, and recent work explores task-oriented grasping via imitation~\cite{chen2025tool} and robust grasp transfer~\cite{Song2020RobustTG}. While effective for grasping, these methods rely on sparse keypoints or coarse alignment and cannot recover local geometric frames, which SemAnCorr provides through dense surface correspondence without task-specific training.

\section{Problem Statement}\label{sec:statement}

Given two object instances represented as triangular meshes $\mathcal{M}_1 = (\mathcal{V}_1, \mathcal{F}_1)$ and $\mathcal{M}_2 = (\mathcal{V}_2, \mathcal{F}_2)$, where $\mathcal{V}_i \subset \mathbb{R}^3$ denotes the vertex set and $\mathcal{F}_i$ the set of faces, the goal is to compute a dense vertex-level correspondence $f: \mathcal{V}_1 \rightarrow \mathcal{V}_2$ that is both \textbf{semantically consistent}: semantically similar parts correspond to each other, and \textbf{geometrically coherent}: the correspondence is smooth and spans the full object surface. We assume each mesh forms a single connected component, which our functional map formulation (Section~\ref{sec:geo-con-fm}) requires for a well-posed spectral basis. To ground the correspondence in semantic structure, we assume there exists $\alpha$ semantically corresponding region pairs between $\mathcal{M}_1$ and $\mathcal{M}_2$: $\mathcal{K} = \{(k_1^i, k_2^i) \mid i = 1,\dots,\alpha\}$, where $k_1^i \subset \mathcal{V}_1$ and $k_2^i \subset \mathcal{V}_2$ are vertex subsets representing semantically similar parts. These regions serve as high-level \emph{semantic anchors} that constrain $f$:
\begin{align}
    f(k_1^i) \approx k_2^i, \quad \forall\, i = 1, \dots, \alpha,
\end{align}
where the approximation indicates a small error, while $f$ varies smoothly over the surface beyond the anchor regions. Given such a correspondence, any object-relative quantity defined on $\mathcal{M}_1$ can be transferred to $\mathcal{M}_2$ by applying $f$.

\section{Method: Semantic Anchored Correspondence}\label{sec:METHOD}

\begin{figure*}[t]
    \centering
    \includegraphics[width=\textwidth]{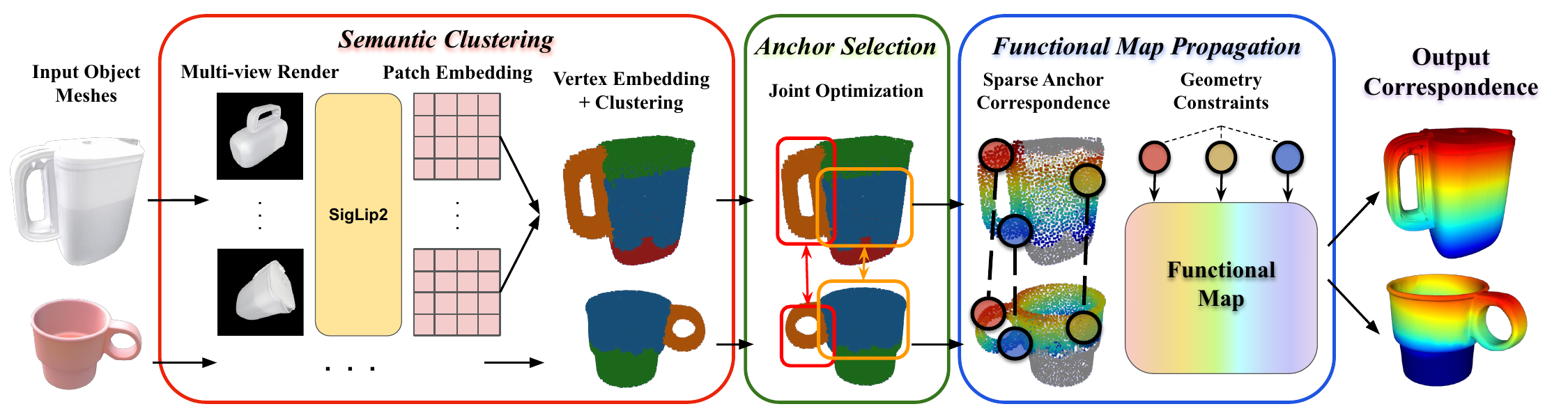}
    \caption{Overview of the proposed SemAnCorr framework. To capture rich semantic representations, SemAnCorr extracts semantic embeddings using pretrained models and lifts them into 3D, and are then used to identify semantic clusters (\cref{sec:semantify}, red box); Corresponding anchor regions are selected through joint semantic-geometric optimization (\cref{sec:sem-anchor}, green box), which constrain a functional map (\cref{sec:geo-con-fm}, blue box) to produce smooth, dense correspondences.}
    \label{fig:methodology-pipeline}
    \vspace{-2em}
\end{figure*}

At a high level, the proposed method first identifies semantically meaningful parts on each object's surface, then selects a small set of corresponding part pairs that are consistent in both meaning and geometry, and finally uses these pairs as constraints to  smoothly propagate a dense correspondence across the object surface. These stages are illustrated in~\cref{fig:methodology-pipeline}. 

\subsection{3D Semantic Acquisition and Clustering}\label{sec:semantify}

Let $\mathcal{M} = (\mathcal{V}, \mathcal{F})$ denote a triangular surface mesh of a 3D object as defined in~\cref{sec:statement}. To obtain multi-view observations of the 3D surface, we define a set of $N$ virtual cameras $\{\mathcal{C}_i\}_{i=1}^N$ and generate RGB images $\mathcal{R}$: $\{ \ \mathbf{I}_i = \mathcal{R}(\mathcal{M};\, \mathcal{C}_i) \mid i = 1, \dots, N\ \}$ using a differentiable renderer, each corresponding to a distinct viewpoint of the mesh.

\textbf{Semantic Feature Extraction.}
For each rendered image $\mathbf{I}_i$, we extract per-patch semantic embeddings using the pretrained SigLip2Vision model~\cite{Zhai2023SigmoidLF}. To obtain richer  representations than a single hidden layer, we average the last $4$ hidden states of the model, yielding
\begin{align}
    \mathbf{E}_i = \Phi(\mathbf{I}_i)
    = \bigl\{\mathbf{e}_{i}^{(p)} \in \mathbb{R}^d\bigr\}_{p=1}^{P},
\end{align}
where $P$ is the number of image patches and $d$ the embedding dimension. Each patch embedding $\mathbf{e}_{i}^{(p)}$ is associated with a 3D surface location via a lifting operator $\mathcal{L}_i : \mathbb{R}^2 \rightarrow \mathbb{R}^3$, defined using camera parameters and depth from the rendering stage. Applying $\mathcal{L}_i$ to each patch and aggregating across all views yields a semantic point cloud $\mathcal{P}_s = \bigl\{(\mathbf{p}_j,\, \mathbf{s}_j)\bigr\}_{j=1}^{M}$, where each point $\mathbf{p}_j \in \mathcal{M}$ is endowed with a
semantic embedding $\mathbf{s}_j \in \mathbb{R}^d$, computed as the
mean of all patch embeddings whose lifted 3D location is nearest to
$\mathbf{p}_j$.

\textbf{Semantic Clustering.}
Given $\mathcal{P}_s$, we partition the object surface into semantically coherent part regions. We first reduce the embedding dimensionality to $d'$ via principal component analysis and form a joint feature vector for each point $\mathbf{x}_j = [\,\hat{\mathbf{s}}_j,\ w_p\,\bar{\mathbf{p}}_j\,]$, where $\hat{\mathbf{s}}_j \in \mathbb{R}^{d'}$ is the reduced semantic embedding, $\bar{\mathbf{p}}_j$ is the normalized 3D position, and $w_p$ controls the relative influence of spatial proximity versus semantic similarity. We then apply $k$-means clustering in this joint feature space to obtain an initial partition of $K$ clusters.

To produce contiguous parts, SemAnCorr refines the initial clusters by splitting spatially disconnected regions into separate parts and merging small fragments into their nearest neighboring cluster. This results in a set of $K$ semantically meaningful parts $\{\mathcal{C}_c\}_{c=1}^K$, each with an associated embedding
\begin{align}
    \bar{\mathbf{s}}_c
    = \frac{1}{|\mathcal{C}_c|}
      \sum_{(\mathbf{p}_j,\mathbf{s}_j)\,\in\,\mathcal{C}_c} \mathbf{s}_j,
\end{align}
which serves as a compact descriptor for downstream computation.

\subsection{Joint Alignment and Anchor Selection}\label{sec:sem-anchor}

Given cluster embeddings $\{\bar{\mathbf{s}}_c^{(1)}\}_{c=1}^{K_1}$ and $\{\bar{\mathbf{s}}_{c'}^{(2)}\}_{c'=1}^{K_2}$ from two object meshes $\mathcal{M}^{(1)}$ and $\mathcal{M}^{(2)}$, our goal is to identify $\alpha$ high-confidence semantic anchor pairs $\mathcal{A} = \{(c, c')\}$ that constrain the downstream functional map.

\textbf{Relative Cosine Similarity.}
Directly comparing raw cluster embeddings conflates object-level category signal with part-discriminative signal: two objects of the same category will have uniformly high cosine similarity across all cluster pairs, making part-level discrimination difficult. To isolate the residual part-discriminative signal from category-level features, we subtract each object's mean embedding prior to comparison:
\begin{align}
    \tilde{\mathbf{s}}_c^{(i)}
    = \frac{\bar{\mathbf{s}}_c^{(i)} - \mu^{(i)}}
           {\|\bar{\mathbf{s}}_c^{(i)} - \mu^{(i)}\|_2 + \epsilon},
    \quad
    \mu^{(i)} = \frac{1}{K_i}\sum_c \bar{\mathbf{s}}_c^{(i)},
\end{align}
and compute the relative similarity matrix
$\mathbf{S} \in \mathbb{R}^{K_1 \times K_2}$ as
\begin{align}
    \mathbf{S}_{cc'} =
    \langle \tilde{\mathbf{s}}_c^{(1)},\,
            \tilde{\mathbf{s}}_{c'}^{(2)} \rangle.
\end{align}

\textbf{Bilateral Margin Anchor Selection.}
We rank anchor pair candidates
by \emph{bilateral margin confidence}, a measure of mutual exclusivity that captures how strongly each cluster prefers its proposed match over all alternatives:
\begin{align}
    m_{\mathrm{row}}(c, c') &=
        \mathbf{S}_{cc'} - \max_{k \in \{1,...,K_2\},\ k \neq c'} \mathbf{S}_{ck}, \\
    m_{\mathrm{col}}(c, c') &=
        \mathbf{S}_{cc'} - \max_{k \in \{1,...,K_1\},\ k \neq c}  \mathbf{S}_{kc'}, \\
    \mathrm{conf}(c, c')    &=
        \min\bigl(m_{\mathrm{row}}(c,c'),\;
                    m_{\mathrm{col}}(c,c')\bigr).
\end{align}
Here, $m_{\mathrm{row}}(c, c')$ measures how much cluster $c$ on $\mathcal{M}^{(1)}$ prefers $c'$ over any other cluster on $\mathcal{M}^{(2)}$, and $m_{\mathrm{col}}(c, c')$ measures the symmetric preference from $c'$'s perspective. Taking the minimum enforces that both sides of the pair are mutually exclusive. We greedily select the $\alpha$ highest-confidence pairs subject to a
bijectivity constraint: once a cluster is assigned, it is excluded from further selection, yielding an anchor set
\begin{align}
    \mathcal{A} \subset \{1,\dots,K_1\} \times \{1,\dots,K_2\},
    \quad |\mathcal{A}| = \alpha,
\end{align}
such that any distinct $(c_1, c_1'), (c_2, c_2') \in \mathcal{A}$ satisfy $c_1 \neq c_2$ and $c_1' \neq c_2'$.

\textbf{Joint Pose-Correspondence Optimization.}
Initial anchor pairs from semantic similarity alone may be geometrically inconsistent: semantically similar clusters can correspond to spatially incompatible regions when objects differ significantly in geometry. We therefore refine the anchor correspondence by jointly optimizing a rigid alignment $(\mathbf{R}, \mathbf{t})$ between the two objects and a soft cluster correspondence. To factor out the global scale, we first normalize both objects to the unit sphere. We then define a joint score for each candidate pair $(c, c')$:
\begin{align}
    \mathrm{Score}(c, c';\, \mathbf{R}, \mathbf{t})
    = \lambda_g \cdot \mathrm{geo}(c, c';\, \mathbf{R}, \mathbf{t})
    + \lambda_s \cdot \mathbf{S}_{cc'},
    \label{eq:sem-anchor-score}
\end{align}
where $\mathrm{geo}(c, c';\, \mathbf{R}, \mathbf{t})$ measures geometric compatibility between cluster $c$ (after applying $\mathbf{R}, \mathbf{t}$) and cluster $c'$, computed as a chamfer-based similarity between cluster point clouds:
\begin{equation}
    \mathrm{geo}(c, c'; \mathbf{R}, \mathbf{t})
    = \exp\bigl(-d_{\mathrm{Ch}}(
        \mathbf{R}\mathcal{P}_c^{(1)} + \mathbf{t},\,
        \mathcal{P}_{c'}^{(2)})
    / 2\sigma^2\bigr),
\end{equation}
where $\mathcal{P}_c^{(i)}$ denotes points from cluster $c$ on mesh $\mathcal{M}^{(i)}$, and $d_{\mathrm{Ch}}$ is the symmetric chamfer distance.

The pose is optimized against a soft-assignment objective in which each cluster on $\mathcal{M}^{(1)}$ distributes its correspondence over clusters on $\mathcal{M}^{(2)}$ according to the joint scores:
\begin{align}
    \mathcal{L}(\mathbf{R}, \mathbf{t})
    &= -\sum_{c=1}^{K_1} \sum_{c'=1}^{K_2}
      w_{cc'}\,\mathrm{Score}(c, c';\, \mathbf{R}, \mathbf{t}), \\
    w_{cc'} &= \operatorname*{softmax}_{c'}
      \bigl(\mathrm{Score}(c, \cdot;\, \mathbf{R}, \mathbf{t})\bigr).
    \label{eq:joint-loss}
\end{align}
The soft weights $w_{cc'}$ act as an implicit correspondence that is co-updated with the pose: the semantic term steers which cluster pairs dominate the geometric alignment gradient, coupling the two even though $\mathbf{S}$ itself is pose-independent. We minimize $\mathcal{L}$ via gradient descent with $B$ initializations, using a cosine schedule such that early iterations weight the assignment toward semantic similarity, pulling the pose to align semantically matched clusters, while later iterations shift weight toward geometric compatibility to refine the alignment. The joint scores of the best initialization $\mathrm{Score}(c, c';\, \mathbf{R}^*, \mathbf{t}^*)$ are the anchor confidence scores that are used for final anchor selection via bilateral margin confidence.

\subsection{Semantic Anchored Functional Map}\label{sec:geo-con-fm}

We compute the dense vertex-level correspondence using a functional map, a compact linear operator that encodes how functions defined on one surface transform to another. Rather than matching individual points directly, a functional map $\mathbf{C} \in \mathbb{R}^{k\times k}$ operates in a low-dimensional spectral basis formed by the first $k$ Laplace-Beltrami eigenfunctions of each mesh. This spectral representation acts as a smoothness prior: since low-frequency eigenfunctions capture large-scale surface structure, the correspondence is biased toward smooth variation. A dense pointwise map is then recovered from $\mathbf{C}$ by nearest-neighbor search in the induced spectral embedding space. We constrain this functional map using semantic anchors $\mathcal{A}$, ensuring it is grounded in semantically meaningful part structure while remaining spatially smooth across the surface.

\textbf{Anchor Region Correspondence.}
For each anchor pair $(c, c') \in \mathcal{A}$, we find vertex correspondences within the anchor regions using the global alignment $(\mathbf{R}^*, \mathbf{t}^*)$ to pre-register the meshes. Within each region, we apply per-anchor normalization, so that each region is centered at its bounding-box center and scaled per-axis by its extents. Vertex correspondences are found by nearest-neighbor search in this normalized space, yielding anchor correspondences $(\mathcal{V}_2^{\mathrm{anc}}, \mathcal{V}_1^{\mathrm{anc}})$.

\textbf{Functional Map Initialization and Refinement.}
We fit an initial functional map $\mathbf{C} \in \mathbb{R}^{k\times k}$ from a sparse set of $n_{\mathrm{kp}}$ keypoints subsampled per anchor cluster, which avoids over-constraining the spectral basis that collapses the pointwise
map into a Voronoi partition:
\begin{align}
    \mathbf{C} = \arg\min_{\mathbf{C}}
    \left\lVert
        \boldsymbol{\Phi}_{2}[\mathcal{V}_2^{\mathrm{kp}},
        {:}k]\,\mathbf{C}
        -
        \boldsymbol{\Phi}_{1}[\mathcal{V}_1^{\mathrm{kp}}, {:}k]
    \right\rVert_F^2,
\end{align}
where $\boldsymbol{\Phi}_{i}$ denotes the Laplace-Beltrami eigenvector matrix for $\mathcal{M}^{(i)}$. An initial pointwise map is recovered by matching each $\mathcal{M}^{(2)}$ vertex to its nearest neighbor in the spectral embedding space $\boldsymbol{\Phi}_2 \mathbf{C} \approx \boldsymbol{\Phi}_1$. We then refine $\mathbf{C}$ using a geometry-constrained variant of ZoomOut~\cite{melzi2019zoomout}, which progressively increases the basis dimension $k$ while alternating between updating the functional map and the correspondence. At each step, we restrict each vertex to search among the $k_{\mathrm{nb}}$ spatial neighbors of its current match to prevent large jumps, and re-pin interior anchor correspondences after every update to prevent drift. This yields a correspondence that is semantically grounded within anchor regions and geometrically smooth across the surface.

\begin{table*}[t]
    \centering
    \small
    \setlength{\tabcolsep}{3pt}
    \renewcommand{\arraystretch}{1.1}
    \begin{tabular*}{\textwidth}{
        @{\extracolsep{\fill}}
        l
        *{7}{C}
        @{\hspace{5pt}}
        C
        @{}
    }
    \toprule
    \textbf{Method}
    & {\textbf{Scissors}}
    & {\textbf{Pliers}}
    & {\textbf{Eyeglasses}}
    & {\textbf{Knife}}
    & {\textbf{Suitcase}}
    & {\textbf{Bottle}}
    & {\textbf{Kettle}}
    & {\textbf{Average}} \\
    \midrule
    FM-WKS
    & 58.5(10.7) & 63.7(16.1) & 59.1(15.1) & 34.3(28.6)
    & 83.8(25.1) & 62.5(37.1) & 79.7(14.4) & 63.1(15.0) \\
    DenseMatcher
    & 71.5(16.7) & 69.4(17.5) & 70.7(14.8) & 63.3(25.5)
    & 54.6(28.8) & 54.9(27.8) & 71.5(7.0)  & 65.1(7.1) \\
    Robo-ABC
    & 72.9(9.1)  & 74.4(10.9) & 62.3(18.5) & 51.9(26.9)
    & 71.5(14.8) & 46.9(26.4) & 62.6(6.8)  & 63.2(9.9) \\
    D3Fields
    & 86.2(4.3)  & 88.9(7.0)  & 85.4(7.8)  & 85.8(7.9)
    & 83.2(13.3) & 78.9(7.4)  & 84.1(8.1)  & 84.6(2.9) \\
    \bfseries SemAnCorr (Ours)
    & \multicolumn{1}{B}{88.5(3.0)}
    & \multicolumn{1}{B}{94.0(2.5)}
    & \multicolumn{1}{B}{93.5(2.3)}
    & \multicolumn{1}{B}{90.8(3.8)}
    & \multicolumn{1}{B}{92.2(5.5)}
    & \multicolumn{1}{B}{89.4(4.5)}
    & \multicolumn{1}{B}{86.9(9.9)}
    & \multicolumn{1}{B}{90.8(2.5)} \\
    \bottomrule
    \end{tabular*}
    \caption{Within-category semantic accuracy across object categories (sAcc $\pm$ Std, \%).}
    \label{tab:inner-cat-acc-results}
    \vspace{-2.0em}
\end{table*}

\section{Empirical Evaluations}\label{sec:EXPERIMENT}

We evaluate SemAnCorr on both a manipulation dataset and real-world robotic settings to answer two questions: 
\begin{enumerate}
    \vspace{-0.2em}
    \item \emph{Can dense correspondence simultaneously preserve semantic consistency and geometric coherence across geometrically diverse object instances?}
    \item \emph{Does improved correspondence quality translate into more reliable zero-shot manipulation skill transfer?}
\end{enumerate}
Throughout, correspondences are visualized by transferring a smooth color gradient from the source object to the target: smooth transitions indicate geometrically coherent mappings, while fragmented patterns indicate spatial inconsistencies.

\subsection{Dataset and Implementation}

We evaluate SemAnCorr on PartNet-Mobility~\cite{Xiang_2020_SAPIEN}, a large-scale dataset of articulated objects with part-level semantic annotations: each mesh vertex $p \in V$ is associated with a semantic part label $l_p$. We select seven object categories comprising over 100 objects, covering both rigid and articulated objects. For each category, we evaluate all object pairs and report the mean and standard deviation of each metric.

We select key hyperparameters as follows: $K=6$ semantic clusters, $\alpha=3$ semantic anchors, $d' = 32$ PCA dimension, $w_p = 0.3$ spatial weight, and $k=30$ Laplace-Beltrami basis functions. These parameters are fixed across all experiments. On an RTX 4090 GPU, the complete SemAnCorr pipeline computes dense correspondence for an object pair in approximately 6\,s on average, including semantic feature extraction, anchor optimization, and functional map estimation.

\subsection{Evaluation Metrics}

\subsubsection{Semantic Accuracy.}

We first evaluate whether a correspondence preserves part-level semantic meaning through \emph{Semantic Accuracy (sAcc)}. Given a correspondence $f:V_1\rightarrow V_2$, each target vertex $p\in V_1$ is assigned the label $l_{f(p)}$ of its corresponding source vertex and compared against its ground-truth label $l_p$:
\begin{align}
\text{sAcc} = \frac{1}{|P|}\sum_{p \in P} \mathbf{}{1}\left[l_p = l_{f(p)}\right],
\end{align}
where $P$ denotes the target vertices with valid semantic labels. 

\subsubsection{Geometric Coherence.}

Transferring a skill requires recovering a consistent local interaction frame on the target, for example, a set of surface directions that must survive the mapping. Since PartNet-Mobility does not provide ground-truth dense correspondence across instances, we cannot measure local-frame error directly. We instead evaluate two necessary conditions for frame recovery, each ruling out a distinct failure mode.

\textbf{Continuity (Cont).}
Local displacement directions transfer coherently only if neighboring vertices on the source map to neighboring points on the target; otherwise the correspondence is \emph{fragmented}. Let $E_1$ denote the edge set of $\mathcal{M}_1$ and $\bar{e}_2$ the mean edge length of $\mathcal{M}_2$. An edge $(u,v) \in E_1$ is \emph{consistent} if its transferred endpoints lie within $\tau$ target edge lengths of each other ($\tau = 2$ in all experiments):
\begin{align}
\text{Cont} = \frac{1}{|E_1|} \sum_{(u,v) \in E_1}
\mathbb{1}\!\left[\, \big\lVert f(u) - f(v) \big\rVert_2
\le \tau \, \bar{e}_2 \,\right].
\end{align}

\textbf{Coverage (Cov).}
Continuity alone is satisfied trivially by a map that \emph{collapses} many source vertices onto a small target region, on which no frame can be defined. We therefore measure the fraction of target vertices that receive at least one correspondence:
\begin{align}
\text{Cov} = \frac{|\{v' \in V_2 : \exists\, v \in V_1,\, f(v) = v'\}|}{|V_2|}.
\end{align}

\begin{table}[t]
    \centering
    \begin{tabular}{l D F F}
    \toprule
    & {\textbf{Continuity $\uparrow$}} & {\textbf{Coverage $\uparrow$}} & {\textbf{GCS $\uparrow$}} \\
    \midrule
    FM-WKS           & \multicolumn{1}{E}{0.96(0.01)} & 0.003(0.001) & 0.007(0.003) \\
    DenseMatcher     & 0.93(0.07) & 0.01(0.01) & 0.01(0.01) \\
    Robo-ABC         & 0.77(0.11) & 0.07(0.06) & 0.11(0.10) \\
    D3Fields         & 0.61(0.08) & 0.10(0.07) & 0.16(0.09) \\
    SemAnCorr (Ours) & 0.93(0.03) & \multicolumn{1}{G}{0.25(0.11)} & \multicolumn{1}{G}{0.40(0.12)} \\
    \bottomrule
    \end{tabular}%
    \caption{Geometric Coherence Score across categories.}
    \label{tab:gcs-results}
\end{table}

\textbf{Geometric Coherence Score (GCS).}
Since frame recovery fails under either failure mode, we combine the two with their harmonic mean, which is dominated by the weaker term:
\begin{align}
\text{GCS} = \frac{2 \cdot \text{Cont} \cdot \text{Cov}}{\text{Cont} + \text{Cov}}.
\end{align}
Note that, as a local average, GCS is largely insensitive to smooth, low-frequency distortions within a part. This design choice is based on the fact that across geometrically distinct instances, the correct correspondence within a part is not uniquely defined by geometry, and scoring it would require committing to an arbitrary parametrization. GCS therefore measures only properties that every valid correspondence must satisfy. Whether a given correspondence is functionally correct depends on the task, which we evaluate directly through real-world task success (Sec.~\ref{sec:rw}).

\subsection{Baselines}

We compare against 4 representative baselines spanning geometry- and semantics-driven correspondence approaches.

\textbf{FM-WKS}~\cite{Aubry2011TheWK} estimates functional map correspondences using Wave Kernel Signature descriptors, operating purely on intrinsic geometric structure without any semantic information, serving as a lower bound on contribution of semantic features.

\textbf{Robo-ABC}~\cite{ju2025robo} establishes pixel-level semantic correspondence in 2D image space using pretrained visual features. We render 8 viewpoints per mesh, take the highest-confidence 2D match, and lift it to 3D. This represents the class of 2D semantic correspondence methods applied to 3D.

\textbf{D3Fields}~\cite{wang2024d3fields} constructs per-vertex 3D descriptors by lifting DINOv2 features from multi-view renders and performing nearest-neighbor matching in feature space. This baseline isolates the contribution of functional map propagation over nearest-neighbor given the same 3D feature lifting strategy.

\textbf{DenseMatcher}~\cite{zhu2024densematcher} combines SD-DINO features with a learned DiffusionNet refiner and functional maps. We evaluate its ability to generalize on our PartNet-Mobility categories.

\begin{figure}[t]
    \centering

    \begin{subfigure}{\linewidth}
        \centering
        \includegraphics[width=0.9\columnwidth]{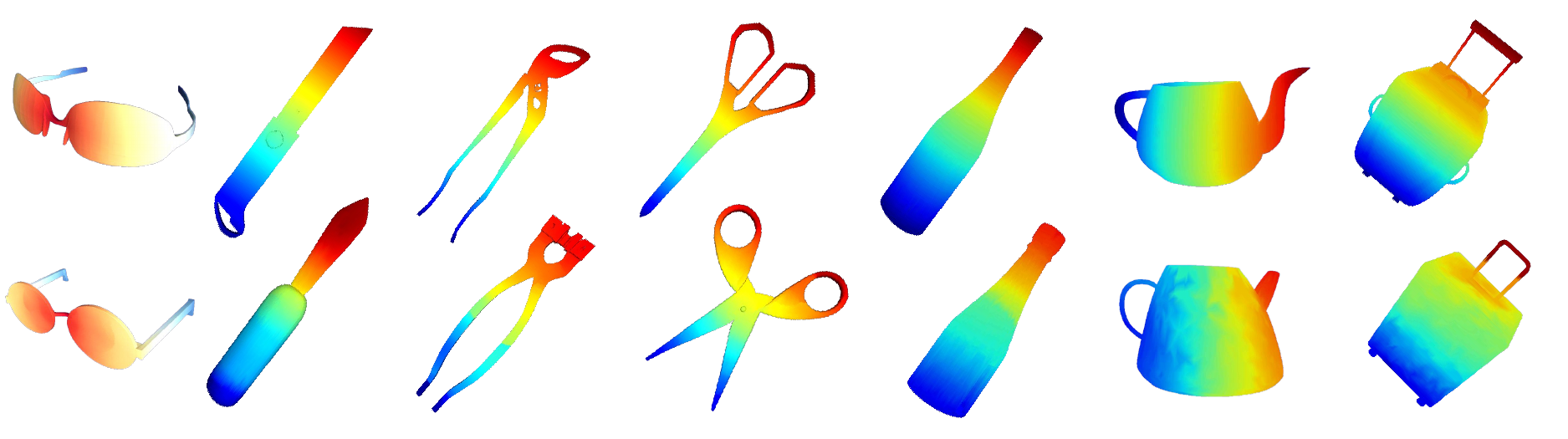}
        \caption{SemAnCorr Inner-Category Correspondence. Top row objects are with source coloring, bottom row objects each vertex takes corresponding point coloring from the source. }
        \label{fig:inner-cat-corr-visual}
    \end{subfigure}

    \begin{subfigure}{\linewidth}
        \centering
        \includegraphics[width=0.9\columnwidth]{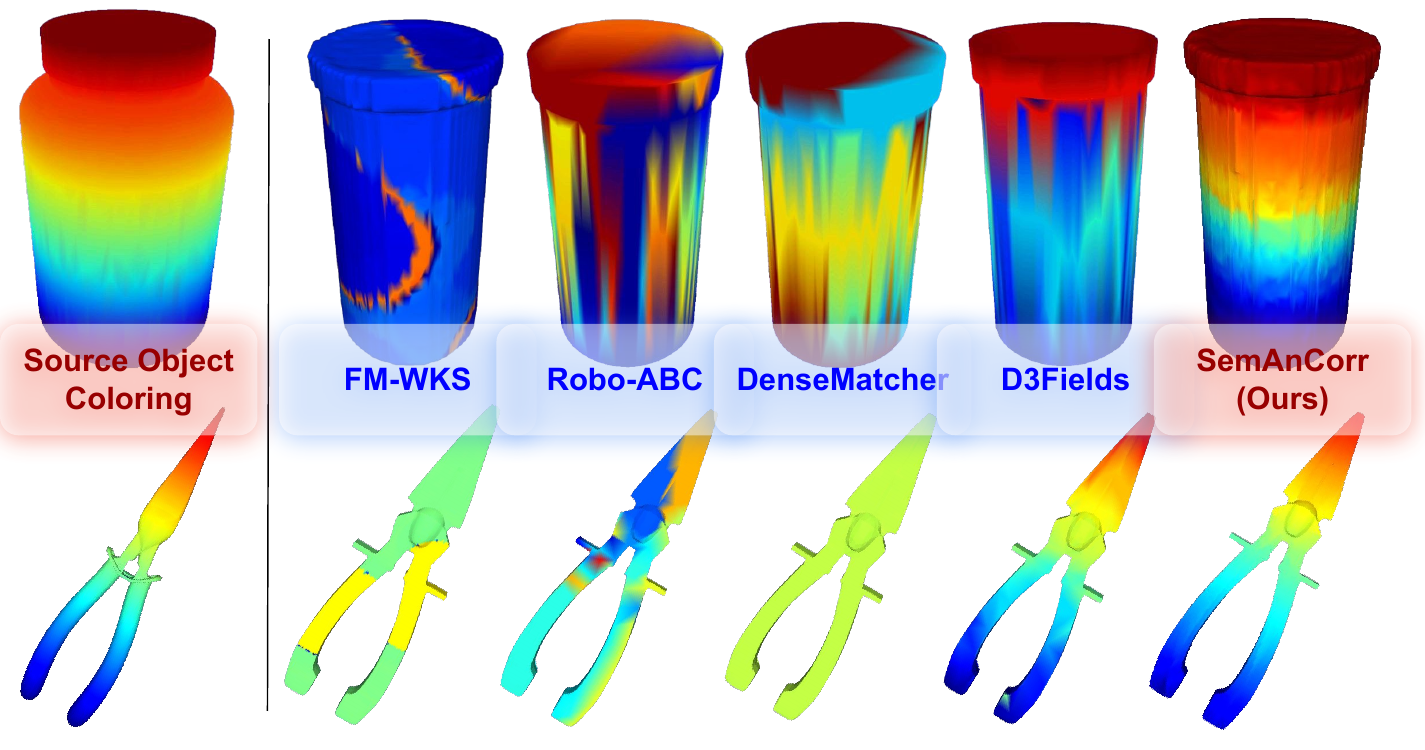}
        \caption{Comparison to Baseline Method Visualization. Source object coloring on the left, corresponded object colorings on the right.}
        \label{fig:baseline-comp-visual}
    \end{subfigure}
    \caption{Inner-Category Correspondence Visualization.}
    \label{fig:inner-cat-corr}
\end{figure}

\subsection{Inner-Category Generalization}\label{sec:inner-cat}

\textbf{Semantic Accuracy.}
We evaluate semantic accuracy for each category and report the results in~\cref{tab:inner-cat-acc-results}. Correspondences are visualized for example object pairs in~\cref{fig:inner-cat-corr-visual} for SemAnCorr, while \cref{fig:baseline-comp-visual} shows comparison across all methods. SemAnCorr outperforms all baselines in all $7$ categories and achieves the highest average accuracy of $90.8\%$, a $6.2$-point improvement over the strongest baseline D3Fields. 

\textbf{Geometric Coherence Score.}
The geometric quality of correspondences is reported in~\cref{tab:gcs-results}. FM-WKS achieves high continuity only by collapsing correspondence onto a small subset of vertices, while D3Fields improves semantic accuracy at the expense of fragmented local mappings. SemAnCorr is the only method that simultaneously achieves high continuity and meaningful surface coverage, resulting in substantially higher GCS, more than twice that of the next best method.

\begin{figure}[t]
    \centering
    \includegraphics[width=0.9\columnwidth]{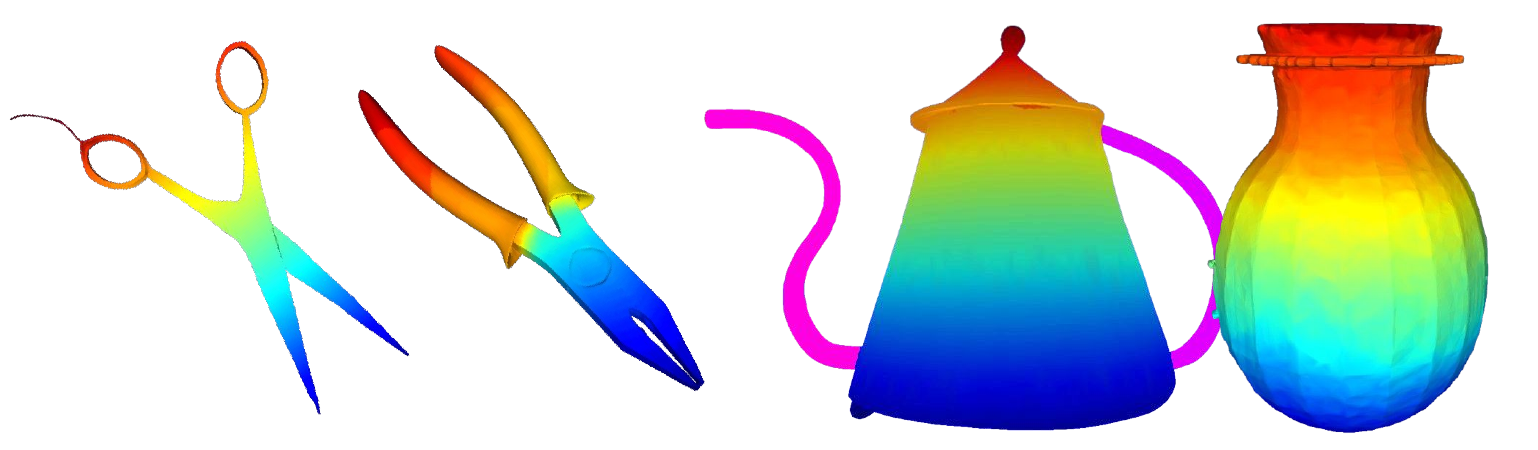}
    \caption{Cross-Category Correspondence Visualization. On the left, Scissors $\rightarrow$ Pliers; On the right, Kettle $\rightarrow$ Bottle.}
    \label{fig:cross-cat-corr-visual}

    \centering
     \resizebox{0.9\columnwidth}{!}{\begin{tabular}{l C C}
    \toprule
    & {\textbf{Scissors $\rightarrow$ Pliers}} & {\textbf{Kettle $\rightarrow$ Bottle}} \\
    \midrule
    FM-WKS           & 55.3(14.7) & 72.3(15.4) \\
    DenseMatcher     & 63.5(18.9) & 65.2(13.8) \\
    Robo-ABC         & 70.3(12.1) & 52.4(15.6) \\
    D3Fields         & 87.2(4.6)  & 83.4(7.5) \\
    SemAnCorr (Ours) & \multicolumn{1}{B}{89.7(2.2)} & \multicolumn{1}{B}{87.8(4.2)} \\
    \bottomrule
    \end{tabular}}
    \captionof{table}{Cross-category semantic accuracy \small (sAcc $\pm$ Std, $\%$).}
    \label{tab:cross-cat-acc-results}
\end{figure}

\begin{figure}[t]
    \centering
    \includegraphics[width=\columnwidth]{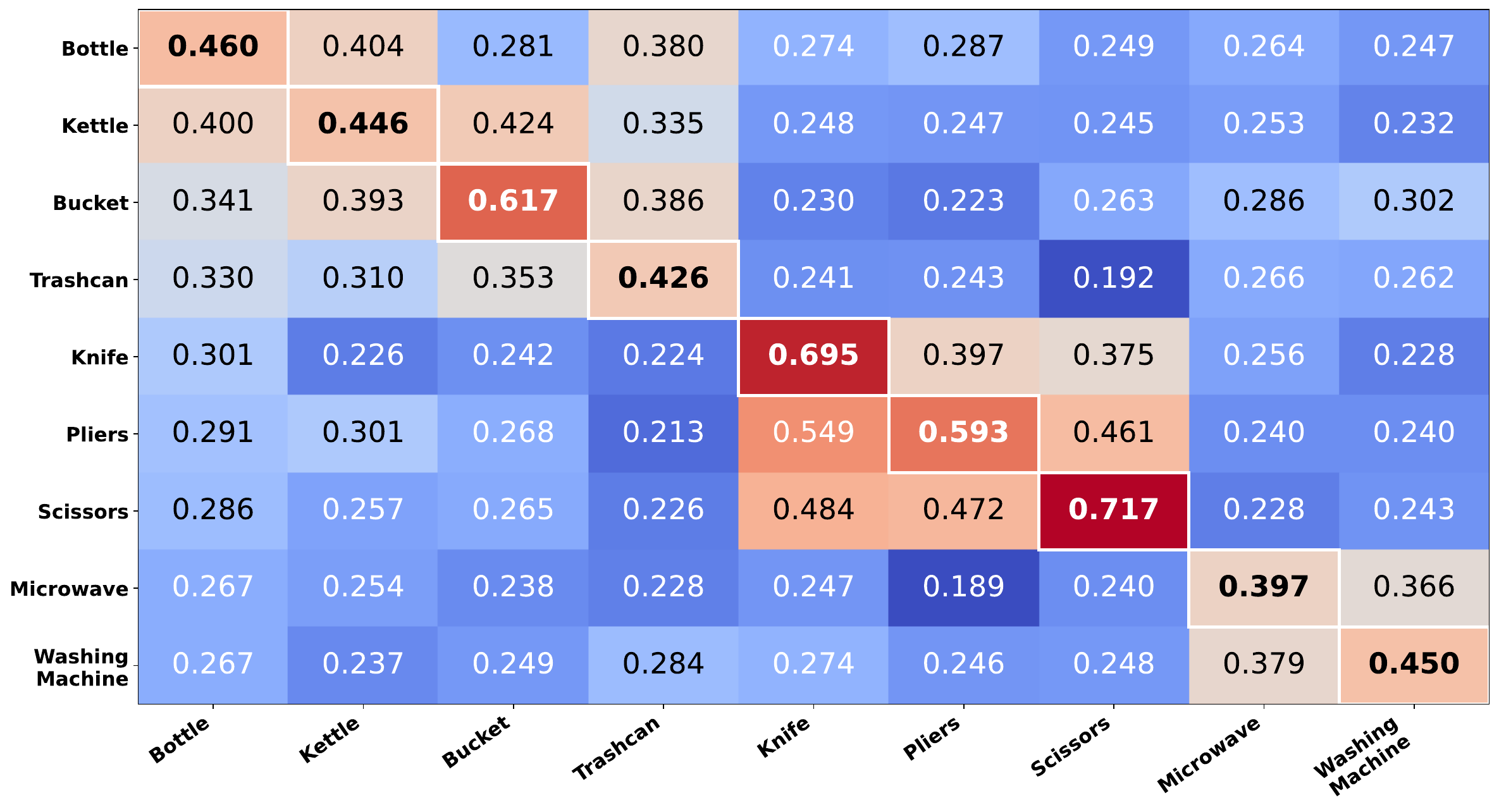}
    \caption{Category-level anchor confidence scores. Within-category pairs (diagonal) score highest, while functionally-related categories score higher than less similar categories.}
    \label{fig:cross-cat-heatmap}
\end{figure}

\subsection{Cross-Category Generalization}

Many manipulation skills naturally transfer across semantically related object categories. We therefore evaluate whether SemAnCorr generalizes beyond category boundaries, and whether the pipeline is able to provide a lightweight signal of cross-category generalization compatibility.

We first evaluate correspondence on two representative cross-category pairs: \emph{Scissors$\rightarrow$Pliers} and \emph{Kettle$\rightarrow$Bottle}. These pairs share functional semantics despite noticeable geometric differences. As shown in~\cref{tab:cross-cat-acc-results} and \cref{fig:cross-cat-corr-visual}, SemAnCorr achieves the highest semantic accuracy on both pairs with smooth visualization, demonstrating that semantic anchor selection generalizes beyond category identity.

We further investigate whether the semantic anchor confidence computed (\cref{sec:sem-anchor}) provides a lightweight signal of compatibility between objects, which may be useful for estimating whether a demonstrated skill is likely to transfer without needing an additional classifier. Since semantic anchor confidence is a novel metric introduced in our method, there is no directly comparable baseline for this analysis. 

\cref{fig:cross-cat-heatmap} shows the average anchor confidence between object categories. The confidence is highest along the diagonal, corresponding to within-category pairs, and remains relatively high among functionally related categories such as containers (bottle, kettle, bucket, trashcan), hand tools (knife, pliers, scissors), and appliances (microwave, washing machine). In contrast, category pairs with less apparent functional or geometric similarity tend to receive lower confidence scores.

These trends suggest that anchor confidence captures a notion of semantic and geometric compatibility rather than category identity alone. While this score is not a direct measure of manipulation success, it may serve as a lightweight prior for deciding generalization compatibility.

\subsection{Ablation Study}

\begin{figure}[t]
    \centering
    \includegraphics[width=0.9\columnwidth]{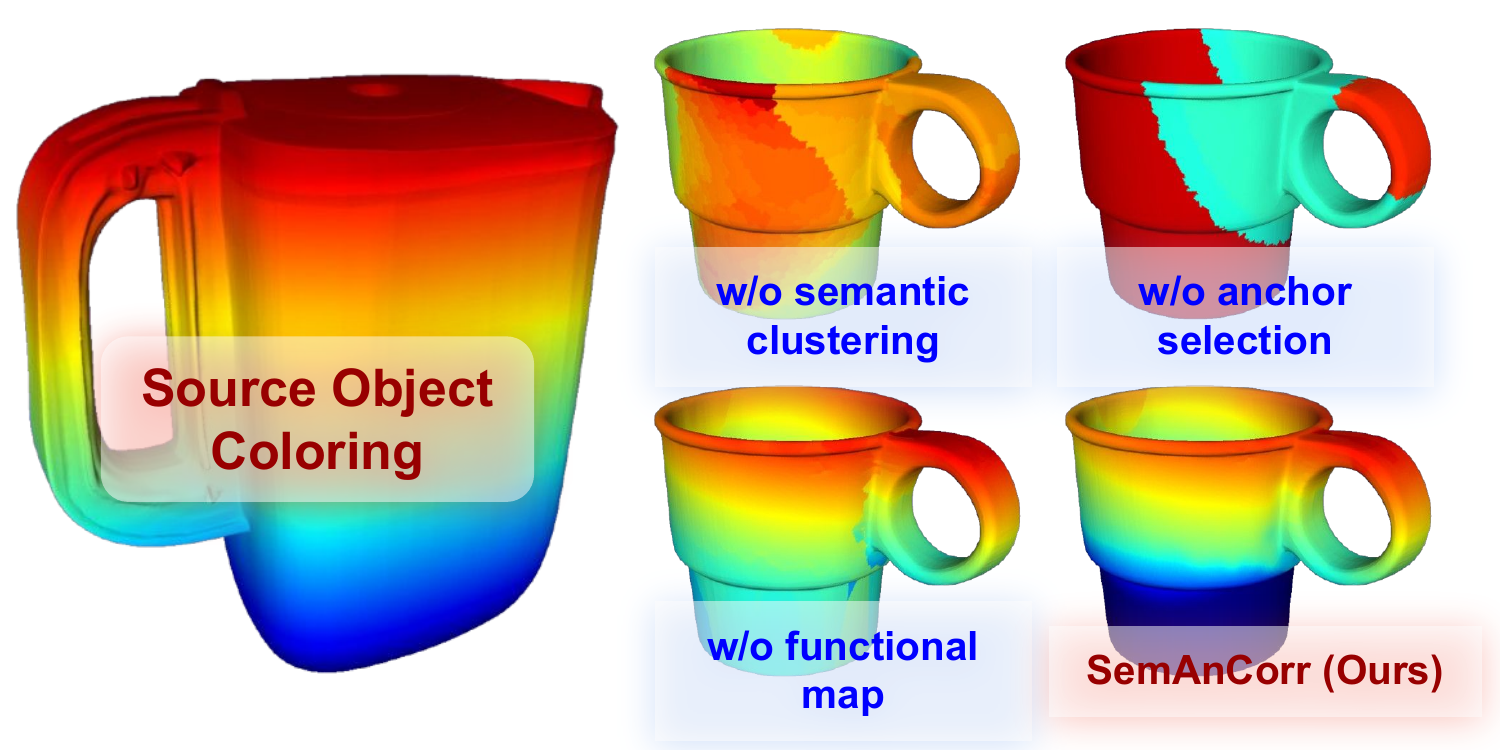}
    \caption{Ablation visualization, with source object (left) and correspondences for each ablation condition (right).}
    \label{fig:ablation-visual}
\end{figure}

\begin{table}[t]
    \centering
    \begin{tabular}{l C D}
    \toprule
    & {\textbf{sAcc}} & {\textbf{GCS}} \\
    \midrule
    SemAnCorr (Ours)        & \multicolumn{1}{B}{90.8(2.5)} & \multicolumn{1}{E}{0.40(0.12)} \\
    w/o Semantic Clustering & 66.0(12.6) & 0.05(0.03) \\
    w/o Anchor Selection    & 39.1(12.1) & 0.08(0.06) \\
    w/o Functional Map      & 72.9(7.4)  & 0.25(0.07) \\
    \bottomrule
    \end{tabular}
    \caption{Ablation study on method components.}
    \label{tab:ablation-results}
\end{table}

We ablate the three principal components of SemAnCorr: semantic clustering, anchor selection, and functional map propagation. The quantitative results are summarized in Table~\ref{tab:ablation-results}, with qualitative examples shown in Fig.~\ref{fig:ablation-visual}. Removing semantic clustering substantially degrades geometric coherence, indicating that clustering suppresses noisy lifted features before correspondence estimation. Removing anchor selection causes the largest drop in semantic accuracy, confirming that semantic anchors provide accurate guidance on how correspondence should be established. Finally, removing functional map propagation reduces both semantic and geometric performance. Together, these results show that semantic consistency and geometric coherence arise from complementary components of the proposed pipeline.

\begin{figure*}[t]
    \centering
    \includegraphics[width=\textwidth]{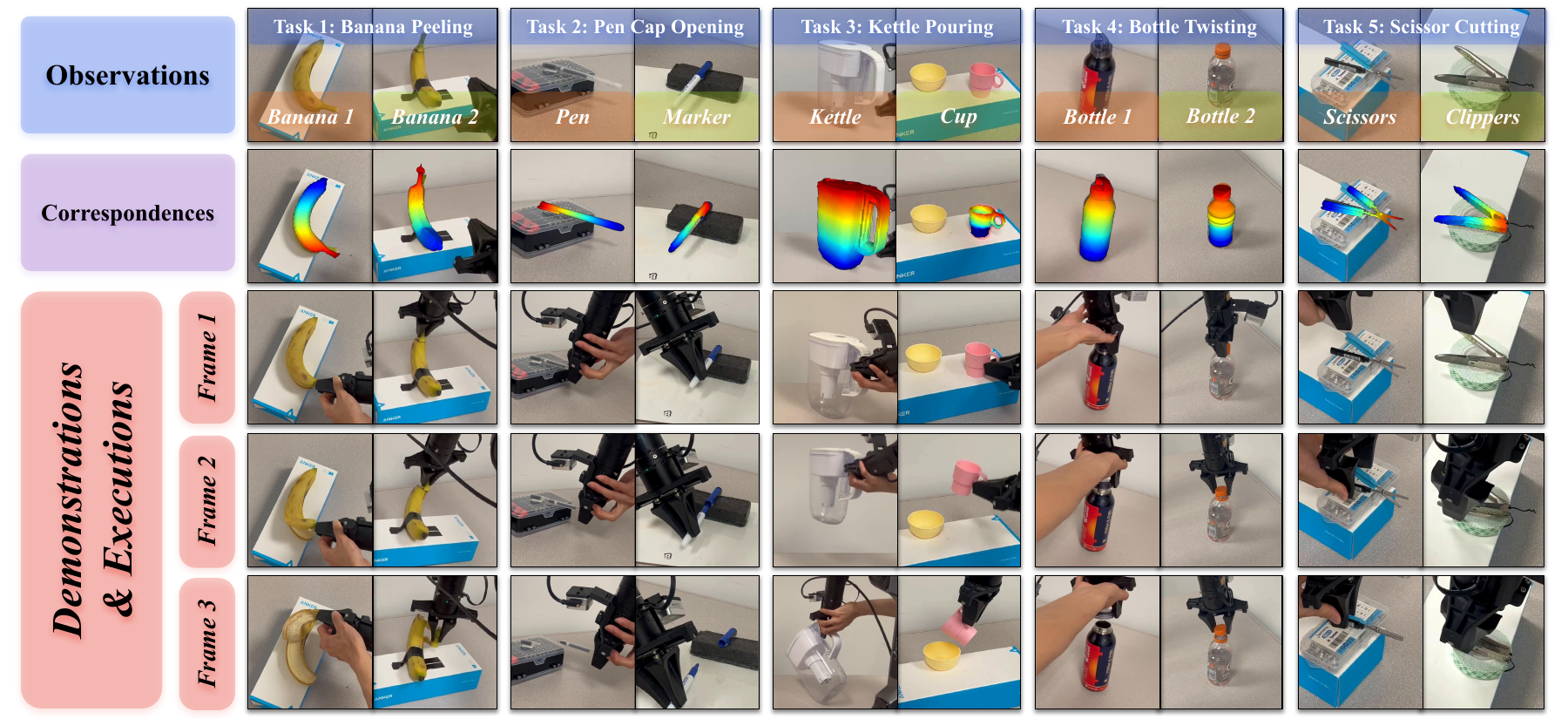}
    \caption{Real-world manipulation visualization. From left to right: Task~\ref{task:banana}, \ref{task:pen-marker}, \ref{task:kettle-cup}, \ref{task:bottle}, and \ref{task:scissor-clipper}. For each task, a manipulation demonstrated on a source object (left) is transferred to a target object (right) using the calculated dense correspondence, visualized in the second row; The remaining rows show key frames of the source demonstration and the transferred execution.}
    \label{fig:rw-execution-visuals}
    \vspace{-1.0em}
\end{figure*}

\begin{figure}[t]
    \centering
    \vspace{-1em}
    \includegraphics[width=\columnwidth]{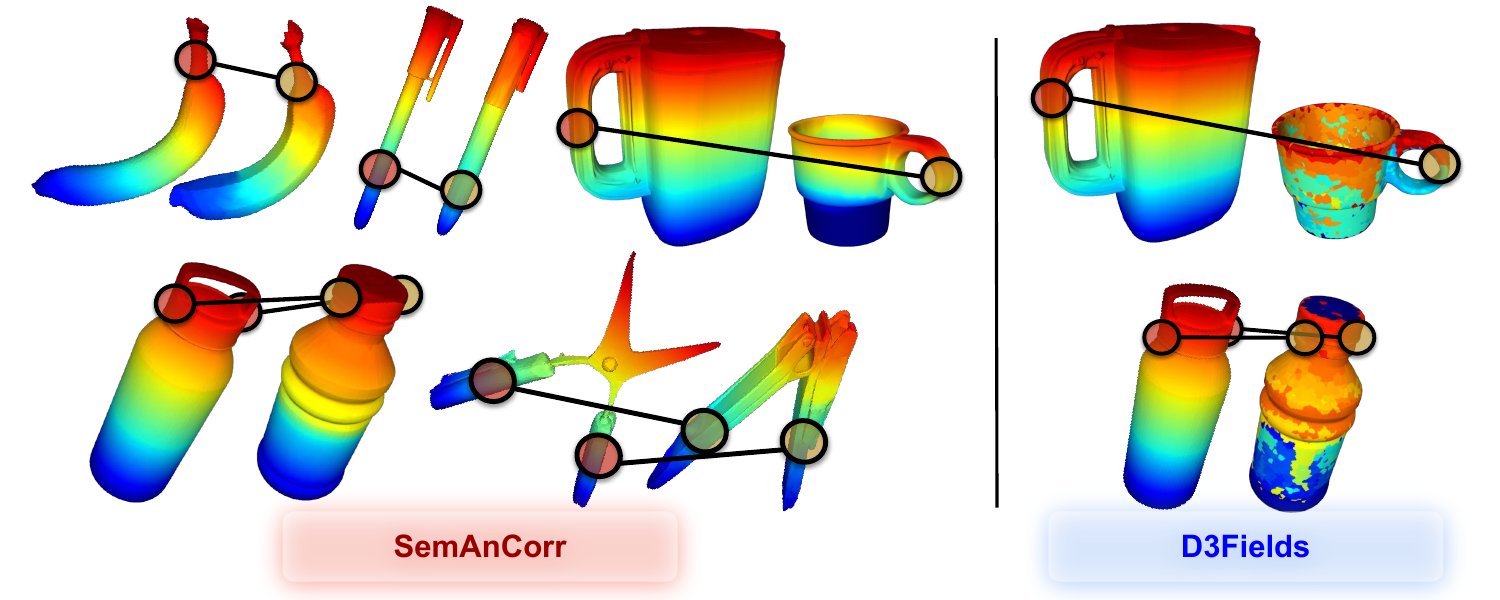}
    \caption{Real-world correspondences. Left: SemAnCorr on Tasks 1–5 (top-left to bottom-right, respectively). Right: D3Fields on Tasks 3 and 4. D3Fields finds the correct semantic parts (circles) but produces spatially incoherent correspondences, yielding incorrect local frames and task failure.}
    \label{fig:rw-corr-visual}
\end{figure}

\subsection{Real-World Manipulation Skill Transfer}\label{sec:rw}

We further deployed SemAnCorr in the real world to test whether the correspondence quality translates into manipulation success. Since generalization across many instances within a category is established in~\cref{sec:inner-cat}, each real-world task transfers a single demonstration to only one unseen target instance, focusing on downstream execution rather than instance breadth. We compare primarily against D3Fields, which is close to SemAnCorr in semantic accuracy but substantially weaker in geometric coherence, isolating the effect of coherence on execution. We select five tasks:
\begin{enumerate}
    \item banana peeling skill to another banana,\label{task:banana}
    \item pen opening skill to a marker,\label{task:pen-marker}
    \item kettle pouring skill to a toy cup,\label{task:kettle-cup}
    \item water bottle twisting opening skill to another bottle,\label{task:bottle} and
    \item scissor cutting skill to a nail clipper.\label{task:scissor-clipper}
\end{enumerate}

We obtain workspace observations using a hand-eye calibrated RGB-D camera, and acquire object mesh using SAM3D~\cite{sam3dteam2025sam3d3dfyimages} from the observation and align the reconstructed mesh to the observed depth point cloud. This aligned mesh serves as both the correspondence input and the canonical frame for skill transfer. Correspondences are shown in~\cref{fig:rw-corr-visual}.

\begin{table}[t]
    \centering
    \begin{tabular}{lccccc}
    \toprule
     & Task~\ref{task:banana} & Task~\ref{task:pen-marker} & Task~\ref{task:kettle-cup} & Task~\ref{task:bottle} & Task~\ref{task:scissor-clipper} \\
    \midrule
    D3Fields         & 7/10 & \textbf{9/10} & 3/10 & 1/10 & 3/10 \\
    SemAnCorr (Ours) & \textbf{8/10} & \textbf{9/10} & \textbf{7/10} & \textbf{7/10} & \textbf{6/10} \\
    \bottomrule
    \end{tabular}
    \caption{Real-world task success rate (out of 10 trials).}
    \label{tab:real-world-success-rate}
\end{table}

\textbf{Skill Representation.}
Dense surface correspondence enables a natural object-centric encoding of manipulation skills. We adopt a simple and practical one based on contact regions and relative waypoints, which we find sufficient for skill representation. We represent a skill as a sequence of alternating \emph{contact keyframes} and \emph{relative motion segments}. A contact keyframe records the end-effector pose $\mathbf{T}_{\mathrm{eef}} \in SE(3)$ relative to the object frame, and the contact region $\mathcal{R} \subset \mathcal{V}$ (mesh vertices near the gripper). A motion segment records waypoints relative to the preceding contact keyframe pose, making the temporal motion pattern inherited from the demonstration transferable without modification. To acquire the skill, we use kinesthetic teaching: recording end-effector poses and gripper state directly, shown in~\cref{fig:rw-execution-visuals}. This eliminates skill extraction noise as an experimental variable, though executions consequently inherit the slow pace of kinesthetic demonstrations. 

\textbf{Skill Transfer.}
Given dense correspondence from SemAnCorr, each contact keyframe is transferred by first applying Procrustes alignment to the corresponding contact region on the target, yielding a prior end-effector pose  $\mathbf{T}_{\mathrm{prior}}$ that captures the semantically correct approach direction. Rather than executing $\mathbf{T}_{\mathrm{prior}}$ directly, we use it to guide grasp generation: candidate grasps from GPG~\cite{tenpas2017gpd} are sampled within the transferred contact region and ranked by grasp quality and proximity to $\mathbf{T}_{\mathrm{prior}}$ in $SE(3)$. This allows the correspondence to provide semantic and geometric guidance while GPG ensures physical feasibility. Motion segments transfer without modification as they are relative to the executed grasp pose.

\textbf{Experimental Results.}
We run each task across 10 trials that vary object poses and viewpoints. We define task success based on whether the task is completed, and report task success rates across 10 trials in~\cref{tab:real-world-success-rate}. Robot motion frames are shown in~\cref{fig:rw-execution-visuals}. Both methods perform comparably on Tasks~\ref{task:banana} and~\ref{task:pen-marker}, where interactions are relatively simple and correspondence is easy to establish. On Tasks~\ref{task:kettle-cup},~\ref{task:bottle}, and~\ref{task:scissor-clipper}, interactions become complex and fine-grained correspondence is crucial for success. Although D3Fields frequently identifies the correct semantic part, incoherent correspondences produce incorrect local frames, leading to failed executions (Fig.~\ref{fig:rw-corr-visual}). SemAnCorr's remaining failures arise primarily from mesh-to-workspace alignment errors, not correspondences. The results confirm semantic accuracy alone is insufficient and geometric coherence is equally necessary for skill transfer success, showing SemAnCorr as a manipulation-oriented representation that bridges visual demonstrations and executable robot actions.

\section{Conclusion}

We presented SemAnCorr, a training-free framework for dense correspondence that preserves both semantic consistency and geometric coherence. Our benchmark and real-world experiments show that both are necessary for skill transfer: semantic consistency determines where to interact, while geometric coherence determines the local frame in which to execute the interaction. 

\textbf{Limitations and Future Work.} SemAnCorr requires a single connected mesh; disconnected parts, such as a separately modeled faucet knob, receive no correspondence. The real-world pipeline relies on single-view reconstruction and depth alignment, making it sensitive to viewpoint, incomplete reconstruction, and alignment error, which the global Laplace-Beltrami basis can propagate beyond where they occur. Future work includes evaluating transfer to multiple real-world targets per demonstration and extending the formulation to bimanual and dexterous manipulation.

\bibliographystyle{ieeetr} 
\bibliography{bib}
\end{document}